\documentclass[journal]{IEEEtranTIE}
\newtheorem{remark}{Remark}
\usepackage{graphicx}
\usepackage{cite}
\usepackage{picinpar}
\usepackage{amsmath}
\usepackage{amsfonts}
\usepackage{url}
\usepackage{flushend}
\usepackage[latin1]{inputenc}
\usepackage{colortbl}
\usepackage{soul}
\usepackage{multirow}
\usepackage{pifont}
\usepackage{color}
\usepackage{alltt}
\usepackage[hidelinks]{hyperref}
\usepackage{enumerate}
\usepackage{siunitx}
\usepackage{breakurl}
\usepackage{epstopdf}
\usepackage{pbox}
\usepackage[table,xcdraw]{xcolor}
\usepackage{makecell}
\usepackage{booktabs}
\usepackage{array,tabularx}
\usepackage{subcaption}
\usepackage{caption}
\usepackage{enumitem}
\usepackage{algorithm}
\usepackage{algorithmic}
\floatname{algorithm}{Procedure}

\usepackage[most]{tcolorbox}

\begin{document}
\title{A General One-Shot Multimodal Active Perception Framework for Robotic Manipulation: Learning to Predict Optimal Viewpoint}

\author{
	\vskip 1em
	
	Deyun~Qin,
	Zezhi~Liu,
    Hanqian~Luo,
    Xiao~Liang,
    and Yongchun~Fang, \emph{Senior~Member,~IEEE}

	\thanks{
	
        This work is supported in part by the National Natural Science Foundation of China under Grant U25A20473 and in part by the National Natural Science Foundation of China under Grant 62233011. (Corresponding author: Yongchun Fang)
		
        Deyun Qin, Zezhi Liu, Xiao Liang, and Yongchun Fang are with the Institute of Robotics and Automatic Information Systems, College of Artificial Intelligence, Nankai University, Tianjin 300350, China, and also with the Tianjin Key Laboratory of Intelligent Robotics, Nankai University, Tianjin 300350, China (e-mail: qindy@mail.nankai.edu.cn, zezhi.liu@mail.nankai.edu.cn, liangx@nankai.edu.cn, fangyc@nankai.edu.cn).
		
		Hanqian Luo is with the College of Artificial Intelligence, Nankai University, Tianjin 300350, China, and also with the Department of Computing, The Hong Kong Polytechnic University, Hong Kong, China (e-mail: hanqian.luo@connect.polyu.hk).
	}
}

\maketitle
	
\begin{abstract}
Active perception in vision-based robotic manipulation aims to move the camera toward more informative observation viewpoints, thereby providing high-quality perceptual inputs for downstream tasks.
Most existing active perception methods rely on iterative optimization, leading to high time and motion costs, and are tightly coupled with task-specific objectives, which limits their transferability.
In this paper, we propose a general one-shot multimodal active perception framework for robotic manipulation. The framework enables direct inference of optimal viewpoints and comprises a data collection pipeline and an optimal viewpoint prediction network.
Specifically, the framework decouples viewpoint quality evaluation from the overall architecture, supporting heterogeneous task requirements. Optimal viewpoints are defined through systematic sampling and evaluation of candidate viewpoints, after which large-scale training datasets are constructed via domain randomization.
Moreover, a multimodal optimal viewpoint prediction network is developed, leveraging cross-attention to align and fuse multimodal features and directly predict camera pose adjustments.
The proposed framework is instantiated in robotic grasping under viewpoint-constrained environments. 
Experimental results demonstrate that active perception guided by the framework significantly improves grasp success rates.
Notably, real-world evaluations achieve nearly double the grasp success rate and enable seamless sim-to-real transfer without additional fine-tuning, demonstrating the effectiveness of the proposed framework.
The project website can be found at: \href{https://nkrobotlab.github.io/MVPNet/}{https://nkrobotlab.github.io/MVPNet/} 
\end{abstract}


\markboth{}%
{}

\definecolor{limegreen}{rgb}{0.2, 0.8, 0.2}
\definecolor{forestgreen}{rgb}{0.13, 0.55, 0.13}
\definecolor{greenhtml}{rgb}{0.0, 0.5, 0.0}

\section{Introduction}

\IEEEPARstart{P}{erception} serves as the initial and critical stage of the robotic manipulation pipeline \cite{Nguyen_2024_CVPR}. The completeness and discriminability of observations directly impact the stability of downstream processes, including decision-making, planning, and execution, thereby substantially determining the overall task success rate \cite{11247705, 8202166, wu2023information}. Consequently, for vision-based robotic manipulation tasks, strategically selecting viewpoints to acquire informative perceptual inputs actively is crucial.

\begin{figure}[t]\centering
	\includegraphics[width=8.8cm]{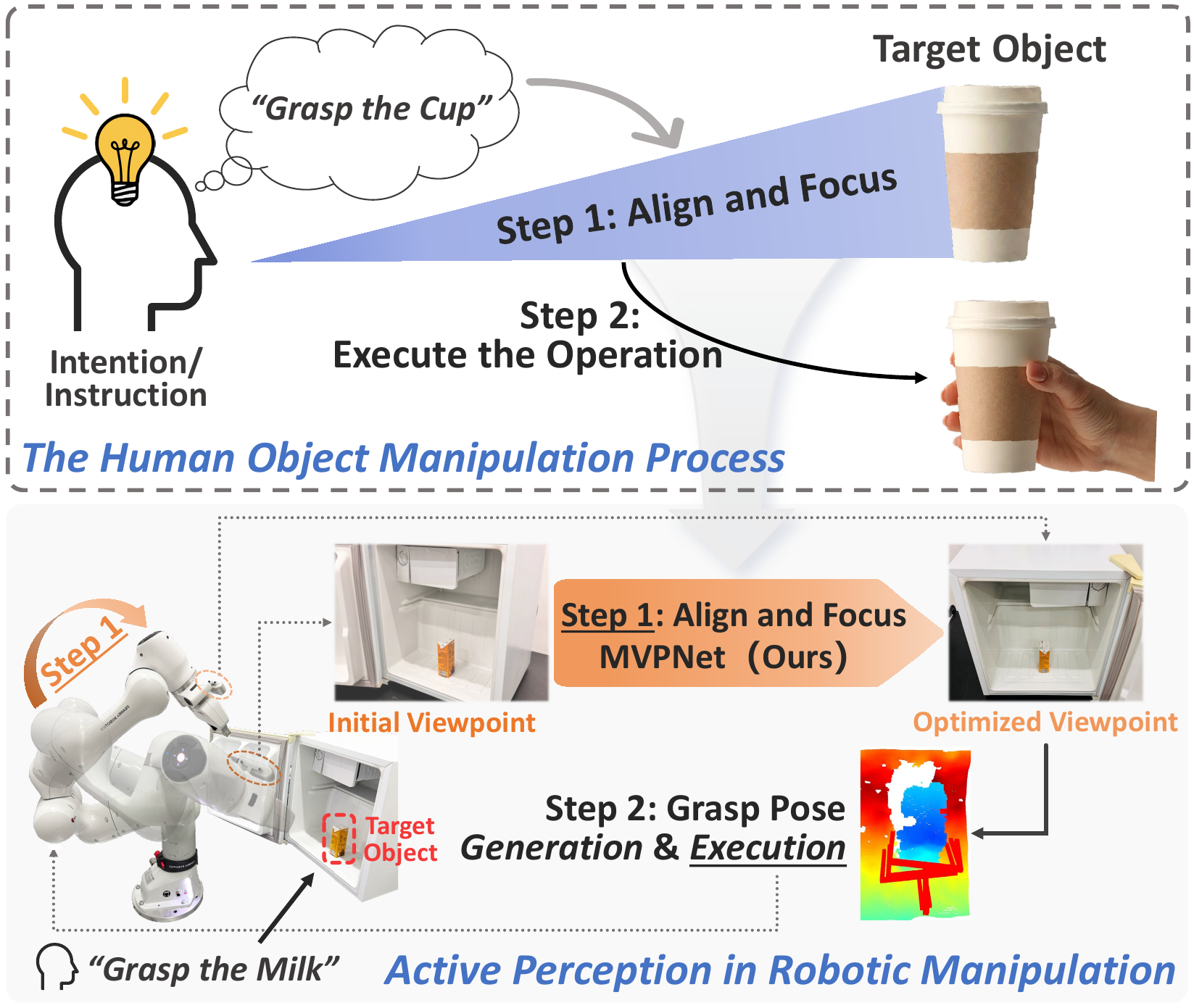}
	\caption{Conceptual illustration of the proposed ``Focus-then-Execute'' active perception paradigm.}
	\label{fig:human_viewpoint_optimize}
\end{figure}

How to actively adjust the camera pose from a random initial viewpoint, based on the current observation and task requirements, to obtain a more informative observation viewpoint constitutes a core problem in active perception research \cite{bajcsy2018revisiting}. In recent years, this problem has attracted extensive attention in various robotic vision tasks, such as robotic grasping \cite{jiang2021synergies, breyer2022closed, zhang2023affordance, ma2024active}, object pose estimation \cite{9826379}, and human pose estimation \cite{Kiciroglu_2020_CVPR, 10801780}. However, most existing methods formulate viewpoint selection as a sequential optimization process, which typically requires multiple perception-decision-action loops to gradually approach an ideal viewpoint. In real robotic systems, such multi-step viewpoint adjustment not only significantly increases time consumption and motion cost, but also imposes higher requirements on environmental reachability and execution stability. Furthermore, existing active perception frameworks are often tightly coupled with specific task objectives and evaluation functions, making them difficult to transfer and reuse across different tasks, thereby limiting their generality.

Achieving effective active perception via a single viewpoint adjustment remains a challenging problem. To address this, we draw inspiration from how humans perform daily tasks. When a task intention or instruction arises, humans typically first reposition and orient themselves toward the target of the task, focusing attention on relevant regions \cite{Bajcsy1985ActivePerception}. This behavior enables them to acquire more comprehensive and discriminative information, which facilitates subsequent actions \cite{11247705, johansson2001eye}, as illustrated in Fig. \ref{fig:human_viewpoint_optimize}. Notably, this adjustment process is typically completed in a single step. This efficiency is attributed to humans' long-term accumulated experience \cite{tarr1989mental, wexler1998motor}, which enables them to directly determine which viewpoints are conducive to perceiving the target object. In contrast, existing robotic systems generally lack this one-shot active perception capability, and a unified framework that can unify the modeling of this capability across diverse task scenarios has yet to be established.

In this paper, we propose a data-driven multimodal active perception framework that directly predicts the optimal observation viewpoint, enabling improved perception with only a single relook, and generalizes across different tasks. Specifically, we first construct a pipeline for optimal viewpoint definition and data collection. The optimal observation viewpoint is determined by sampling candidate viewpoints and evaluating them via a task-dependent quality function, which can be tailored to accommodate diverse task requirements. Subsequently, large-scale training datasets are constructed via domain randomization. Notably, this pipeline eliminates the need for manual annotation, thereby substantially reducing the cost of dataset creation and facilitating rapid adaptation to new tasks. Second, we propose a multimodal optimal observation viewpoint prediction network, referred to as MVPNet. Leveraging the cross-attention mechanism, the proposed network adaptively aligns and fuses multimodal features while highlighting perceptually critical regions, enabling efficient active perception. Finally, we instantiate our framework in robotic grasping under viewpoint-constrained environments, as this task is highly sensitive to observation viewpoints. The main contributions of this work are summarized as follows.

\begin{enumerate}
\item A general one-shot multimodal active perception framework is proposed, comprising a data collection pipeline and an optimal viewpoint prediction network. This framework enables the unified modeling of diverse task requirements, thereby extending its applicability to a broader range of task scenarios.

\item An optimal observation viewpoint data collection pipeline is established, in which optimal viewpoints are defined through task-specific viewpoint quality evaluation functions, and large-scale datasets are constructed via domain randomization.

\item An optimal observation viewpoint prediction network is developed. Utilizing the cross-attention mechanism, this network aligns and fuses multimodal features to predict the required camera pose adjustment.

\item The proposed framework is instantiated in robotic grasping under viewpoint-constrained environments, where data collection and network training are conducted, and its effectiveness and robustness are validated through extensive simulation and real-world experiments.
\end{enumerate}

\section{Related Work}

\subsection{Viewpoint Quality Evaluation}
Currently, there is no unified definition of an ``optimal viewpoint.'' Existing viewpoint theories evaluate viewpoints according to various criteria, including visible area, silhouette, depth, surface curvature, semantics, and aesthetics. Wu \textit{et al.} \cite{11247705} design three task-agnostic viewpoint quality metrics to provide a comprehensive and uniform evaluation of viewpoints.
In contrast, in most mainstream robotic active perception tasks, the notion of an optimal viewpoint is generally regarded as task-dependent. For instance, Ma \textit{et al.} \cite{ma2024active} utilize the distribution of grasp poses to guide camera movements. In contrast, Chen \textit{et al.} \cite{10801780} assess the current human pose to identify the most promising next viewing angles for a drone. Although such task-specific active perception frameworks improve performance on the target task, their viewpoint evaluation functions are tightly coupled with the overall framework, which limits generalization across different task domains.

The framework proposed in this paper decouples the viewpoint quality evaluation function from the overall architecture, allowing task requirements to be accommodated by simply redefining the evaluation function. This design significantly improves the generality and extensibility of the framework.

\subsection{Deep Learning-Based Grasp Pose Estimation}
Compared with traditional analytic methods based on geometric reasoning\cite{sahbani2012overview, bicchi2000robotic, shimoga1996robot}, learning-based approaches for grasp pose estimation have achieved substantial gains in success rate, robustness, and generalization owing to advances in deep learning\cite{newbury2023deep, kleeberger2020survey, kroemer2021review, duan2021robotics, wu2023information, yu2022egnet}.

Depending on the 3D representation used as the network input, existing methods can be broadly categorized into Truncated Signed Distance Field (TSDF)-based methods and point cloud-based methods.
Breyer \textit{et al.} \cite{breyer2021volumetric} propose the Volumetric Grasping Network (VGN), which takes TSDF as input and directly predicts grasp quality, orientation, and width within the voxel space. Furthermore, Yu \textit{et al.} \cite{yu2025trustworthy}, using VGN as the baseline, introduce the trustworthy robotic grasping problem and propose a credibility alignment framework to improve the consistency between predicted grasp probabilities and actual grasp success outcomes. Alternatively, Fang \textit{et al.} \cite{fang2020graspnet} release the large-scale point cloud grasping dataset GraspNet-1Billion and develop the GraspNet grasp pose estimation framework, providing a unified benchmark for general object grasping research. Building on this benchmark, Fang \textit{et al.} \cite{fang2023anygrasp} further propose AnyGrasp, which leverages cross-view geometric consistency and object centroid modeling to improve grasping stability. Wu \textit{et al.} \cite{wu2024economic} present Economic Grasp, implementing more efficient training through an economical supervision strategy that selects critical training samples.

\begin{figure*}[!t]
    \centering
    \includegraphics[width=18.1cm]{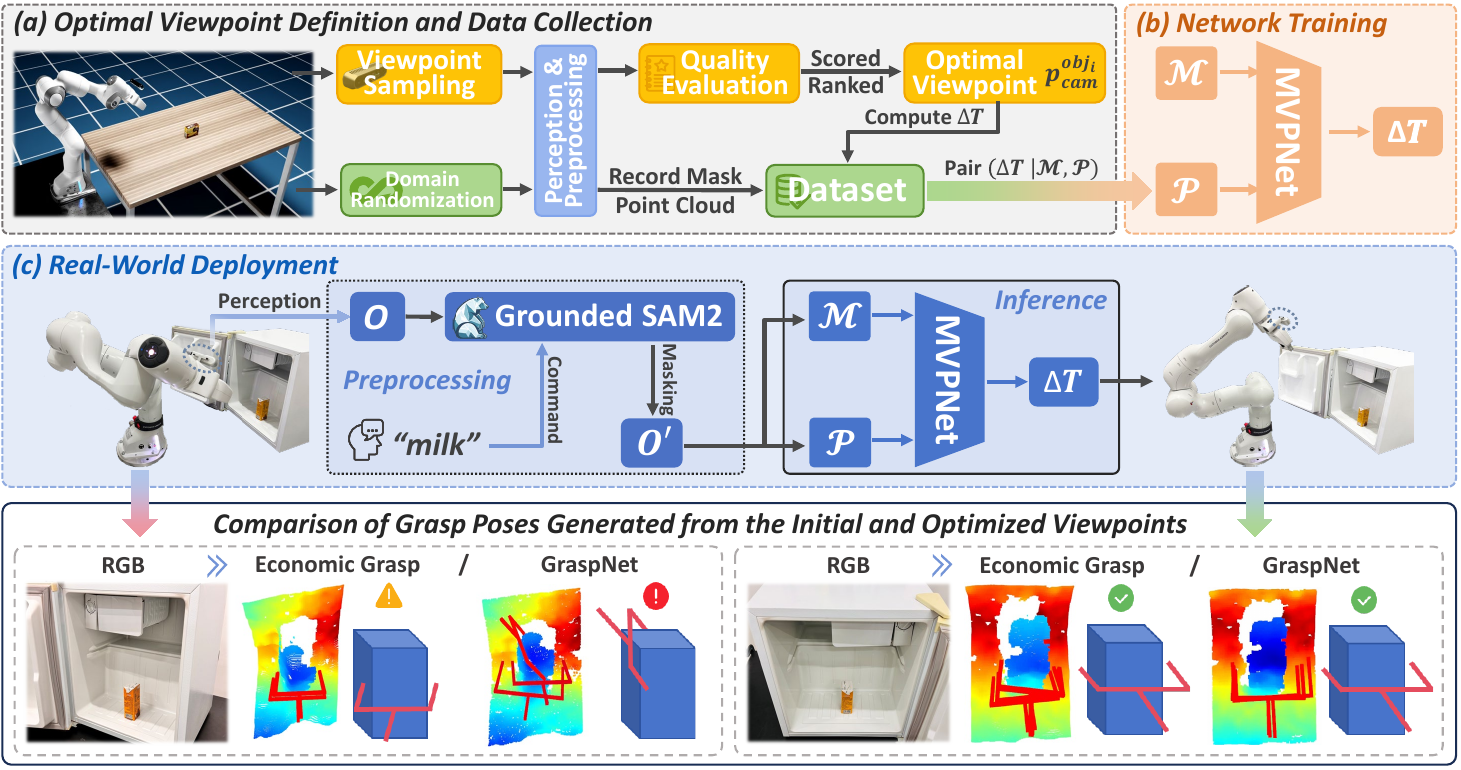}
    \caption{Overall framework of the proposed method, illustrated with robotic grasping in viewpoint-constrained environments:
(a) sampling and evaluating candidate viewpoints to obtain the optimal viewpoint for each object, followed by dataset construction via domain randomization;
(b) training the MVPNet based on the constructed dataset; and
(c) deploying the trained network and conducting comparative evaluations.}
    \label{fig:overall_framework}
\end{figure*}

However, most existing models \cite{fang2020graspnet, Wang_2021_ICCV, fang2023anygrasp, wu2024economic} are designed for tabletop grasping under top-down views. When directly applied to semi-enclosed environments, such as refrigerators or cabinets, where top-down viewpoints are unavailable, and the camera is initialized at a random pose, these models may fail to produce reliable grasp poses due to the significant domain gap between the training data. Such failure cases are illustrated in the bottom-left of Fig.~\ref{fig:overall_framework}. Therefore, in viewpoint-constrained environments, selecting an optimal observation viewpoint to provide more informative inputs for grasp pose estimation models is crucial. Accordingly, this task is chosen to instantiate and validate the proposed framework.

\subsection{Active Perception for Robotic Grasping}
For robotic grasping in viewpoint-constrained environments, active perception enables the acquisition of more informative object observations, thereby better supporting grasp pose estimation. Breyer \textit{et al.} \cite{breyer2022closed} propose a closed-loop Next-Best-View planning strategy driven by occlusion cues, which incrementally updates scene reconstruction and decides online whether to continue observation or execute grasping. Zhang \textit{et al.} \cite{zhang2023affordance} present an affordance-driven Next-Best-View method (ACE-NBV) that, under a view-consistency assumption, uses novel-view synthesis to predict grasp affordance distributions for unobserved viewpoints and selects the next view by maximizing the virtual grasp quality. Ma \textit{et al.} \cite{ma2024active} introduce the Neural Graspness Field, which constructs an online neural field representation of grasp distributions during camera motion and plans next views by targeting regions of high uncertainty. 

These methods typically require multiple optimization steps to reach a desired observation viewpoint, which increases both time and motion costs. Moreover, such methods are tightly coupled with robotic grasping, making them difficult to generalize to other tasks. In contrast, this paper proposes a data-driven active perception framework that can reach an optimal viewpoint in a single adjustment step and can be readily extended to other tasks.

\section{Methodology}
{
The overall active perception framework is designed to learn a policy $\pi: (\mathcal{O}, \mathcal{L}) \mapsto \mathcal{T}$, which takes the current observation $\mathcal{O}$ and the natural language instruction $\mathcal{L}$ specifying the target object as inputs, and outputs the required camera pose adjustment $\mathcal{T}$. The overall implementation consists of three stages:
(a) \textbf{Synthetic Dataset Construction}. Developing a large-scale photorealistic dataset for network training and evaluation;
(b) \textbf{Perception and Preprocessing}. Acquiring environmental observations and performing preprocessing with integrated semantic information;
(c) \textbf{Network Architecture Construction}. Building a transformer-based network that maps the preprocessed multimodal visual information to the required camera pose adjustment.
Here, the framework is instantiated using robotic grasping in viewpoint-constrained environments to validate the overall pipeline.

\subsection{Synthetic Dataset Construction}
Isaac Sim \cite{website1} is leveraged to accelerate dataset construction, harnessing its photorealistic rendering and robust domain randomization for high-fidelity and large-scale synthetic data collection. Moreover, the process does not require manual annotation, thereby reducing the cost of dataset construction.

\begin{figure}[t]
  \centering
  
  \begin{subfigure}[b]{0.677\linewidth}
    \centering
    \includegraphics[width=\linewidth]{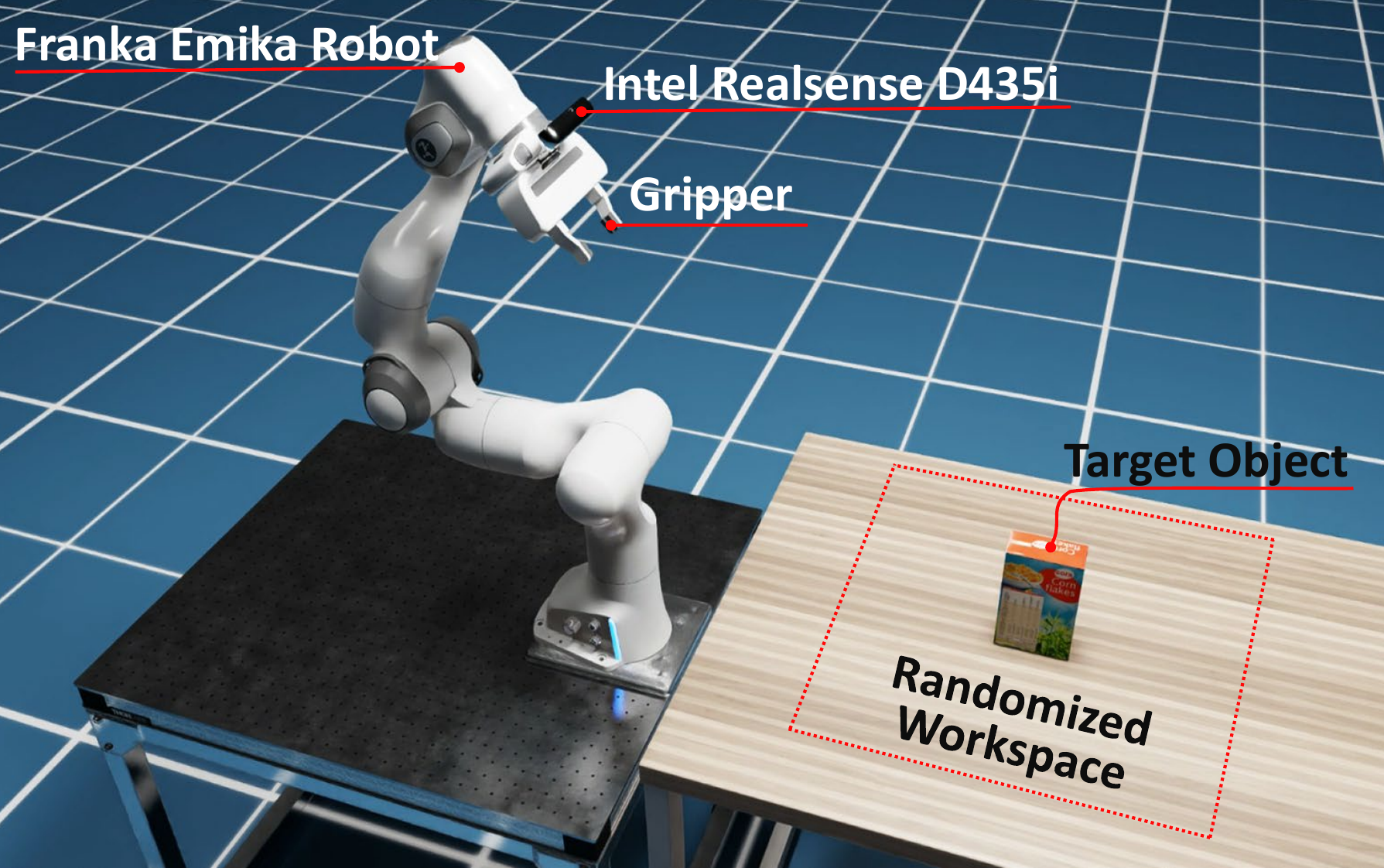}
    \caption{}
    \label{fig:simutation_setup}
  \end{subfigure}
  \hfill
  \begin{subfigure}[b]{0.308\linewidth}
    \centering
    \begin{subfigure}[b]{\linewidth}
      \centering
      \includegraphics[width=\linewidth]{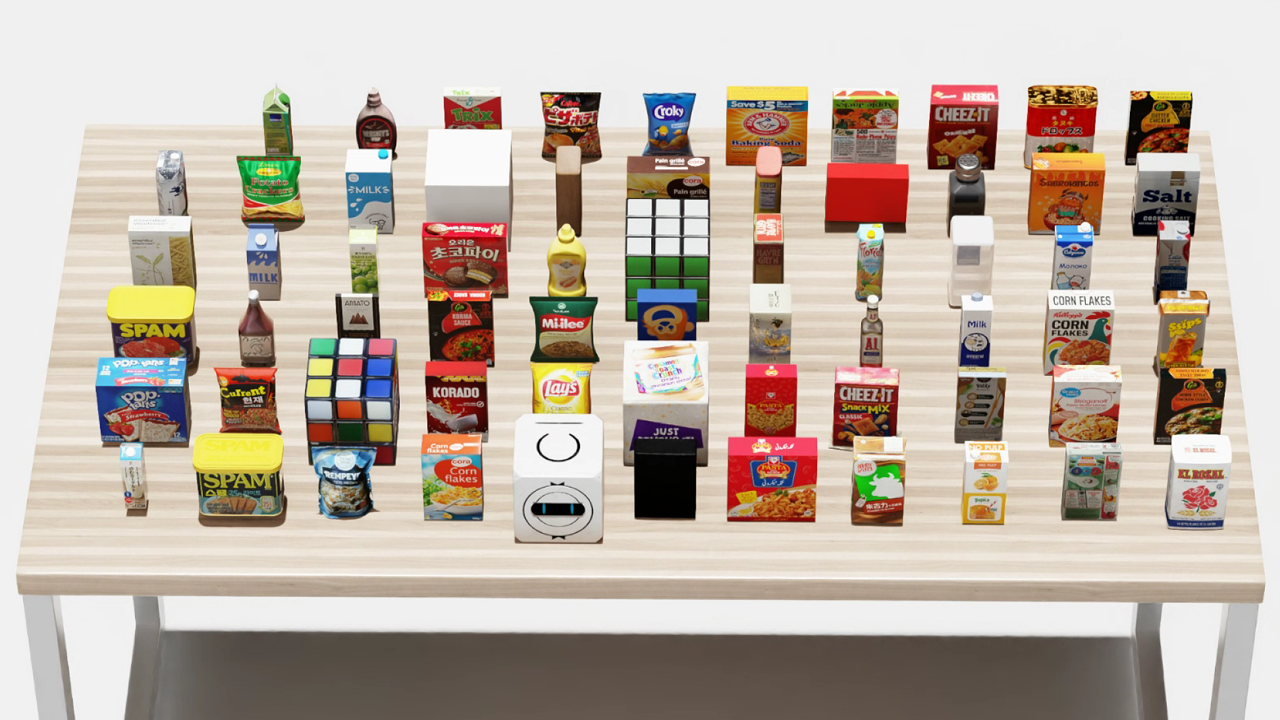}
      \caption{}
      \label{fig:object_similar}
    \end{subfigure}
    \begin{subfigure}[b]{\linewidth}
      \centering
      \includegraphics[width=\linewidth]{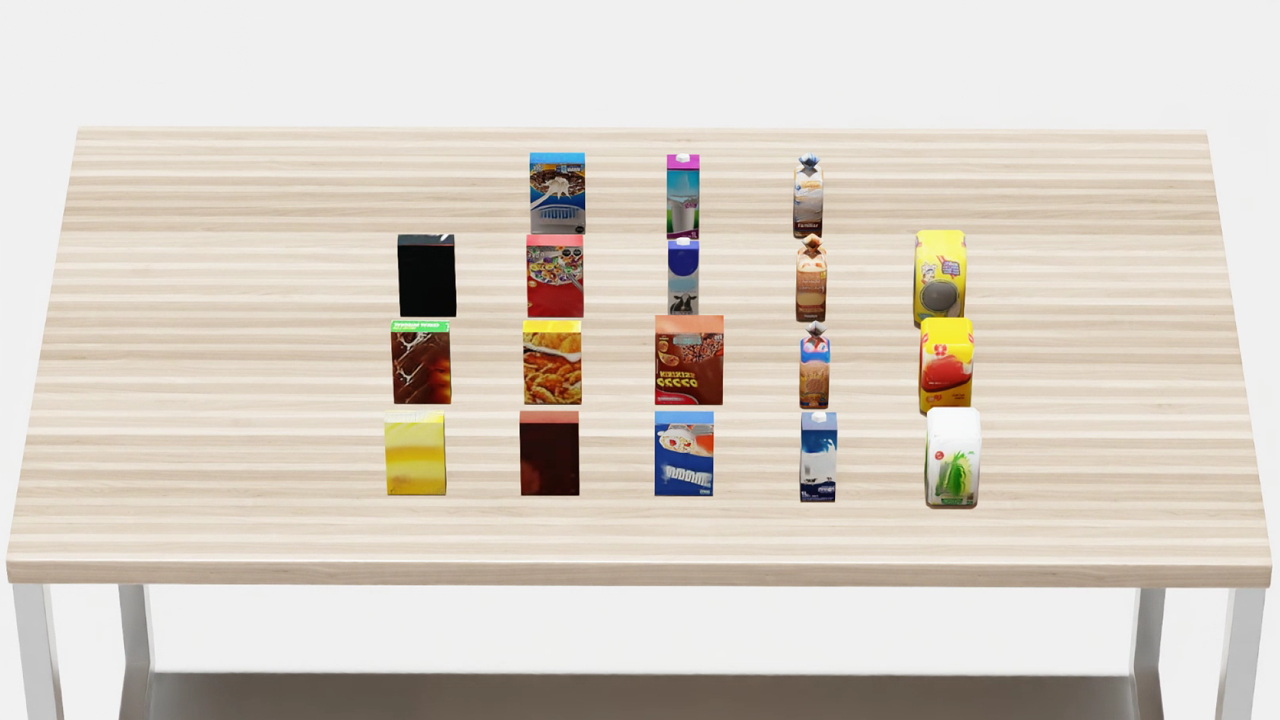}
      \caption{}
      \label{fig:object_novel}
    \end{subfigure}
  \end{subfigure}
  
  \caption{(a) Simulation setup; (b) Similar objects; (c) Novel objects. The simulation environment based on Isaac Sim serves for synthetic dataset construction and simulation-based experimental testing.}
  \label{fig:overall}
\end{figure}

\subsubsection{Define Optimal Observation Viewpoint}
To construct a dataset for network training, we first define the optimal viewpoint for each object as the dataset label by sampling candidate viewpoints around the object and evaluating their quality. Specifically, we construct a simulated scene in Isaac Sim, as shown in \mbox{Fig. \ref{fig:simutation_setup}}, where initial camera viewpoints are randomly sampled around the object, all oriented toward the object center, while RGB-D images and camera poses are recorded. These images, together with the corresponding natural language descriptions of the target objects, are sequentially processed by Grounding DINO \cite{liu2023grounding} and SAM2 \cite{ravi2024sam2segmentimages} for object detection and segmentation. Meanwhile, Economic Grasp \cite{wu2024economic} is employed as the grasping model to compute grasp pose candidates, whose aggregated candidate scores are used to evaluate the quality of each viewpoint. For different downstream tasks, the viewpoint quality evaluation function can be customized accordingly. To ensure the reliability of the process, we randomize 1,500 viewpoints (i.e., camera positions) per object and conduct five repeated grasp pose detections for each point cloud, averaging the top 10 scores in each trial to improve the stability of the evaluation. The process involves 65 object categories, as shown in \mbox{Fig. \ref{fig:object_similar}}. In total, it yields 97.5k images, 487.5k grasp detections, and 4.8 million grasp poses.
As an example, the observation viewpoint score distribution of the object is shown in \mbox{Fig.~\ref{fig:synthetic_dataset_examples}}. Since the scores exhibit continuous distributions in both the 3D space and their 2D projections, this problem can be effectively modeled and optimized in continuous space using neural networks.
The top 800 highest-scoring viewpoints undergo DBSCAN\cite{ester1996density} clustering, with the centroid of the largest cluster selected as the object's optimal observation viewpoint $\mathbf{t}_{best}$ for the current state.

\begin{align}
	\mathbf{t}_{\text{best}} =
	\mathbf{t}_{\text{obj}_i}
	+ \mathbf{R}(\mathbf{q}_{\text{obj}_i}) \cdot
	\mathbf{p}_{\text{cam}}^{\text{obj}_i}.
	\label{eq:t_best}
\end{align}
During viewpoint sampling, the object translation
$\mathbf{t}_{\text{obj}_i}$ and rotation
$\mathbf{R}(\mathbf{q}_{\text{obj}_i})$ remain constant. We define
$\mathbf{p}_{\text{cam}}^{\text{obj}_i}$ to transform the optimal viewpoint
$\mathbf{t}_{\text{best}}$ from the world coordinate frame into the
$\text{obj}_i$ coordinate frame, as shown in
Eq.~\eqref{eq:t_best}. With this formulation,
$\mathbf{t}_{\text{best}}$ can be computed for varying object positions
and orientations.

\subsubsection{Data Collection}
Once the optimal observation viewpoint is obtained for each object, we perform data collection using domain randomization, where initial camera viewpoints, object categories, poses, scales, and environmental lighting are randomized. The network inputs include mask images and point clouds, which are
obtained by preprocessing the current-state observations. The camera pose adjustment
$\Delta \mathbf{T} = (\Delta \mathbf{t}, \Delta \mathbf{q})$ is used as
the learning target of the network and is computed as:
\begin{align}
	\Delta \mathbf{t} = \mathbf{R}(\mathbf{q}_{\text{cam}}^{-1}) \cdot (\mathbf{t}_{\text{best}} - \mathbf{t}_{\text{cam}}), \;
	\Delta \mathbf{q} = \mathbf{q}_{\text{cam}}^{-1} \cdot \mathbf{q}_{\text{target}},
	\label{eq:delta_t and q}
\end{align}
where $\Delta \mathbf{t} \in \mathbb{R}^3$ denotes the translation vector,
and $\Delta \mathbf{q} \in \mathbb{R}^4$ denotes a unit quaternion
satisfying $\|\Delta \mathbf{q}\| = 1$.
The optimal observation viewpoint $\mathbf{t}_{\text{best}}$ is obtained from the current pose of the object $\text{obj}_i$ and its optimal viewpoint position $\mathbf{p}_{\text{cam}}^{\text{obj}_i}$ using \mbox{(\ref{eq:t_best})}. The randomly initialized camera translation $\mathbf{t}_{\text{cam}}$ and rotation $\mathbf{q}_{\text{cam}}$ can be obtained directly from the simulator. The quaternion $\mathbf{q}_{\text{target}}$ represents an orientation that faces the direction of the vector $\mathbf{v} = \mathbf{t}_{\text{best}} - \mathbf{t}_{\text{obj}}$.

\begin{figure}[t]\centering
	\subfloat[]{\includegraphics[width=4.25cm]{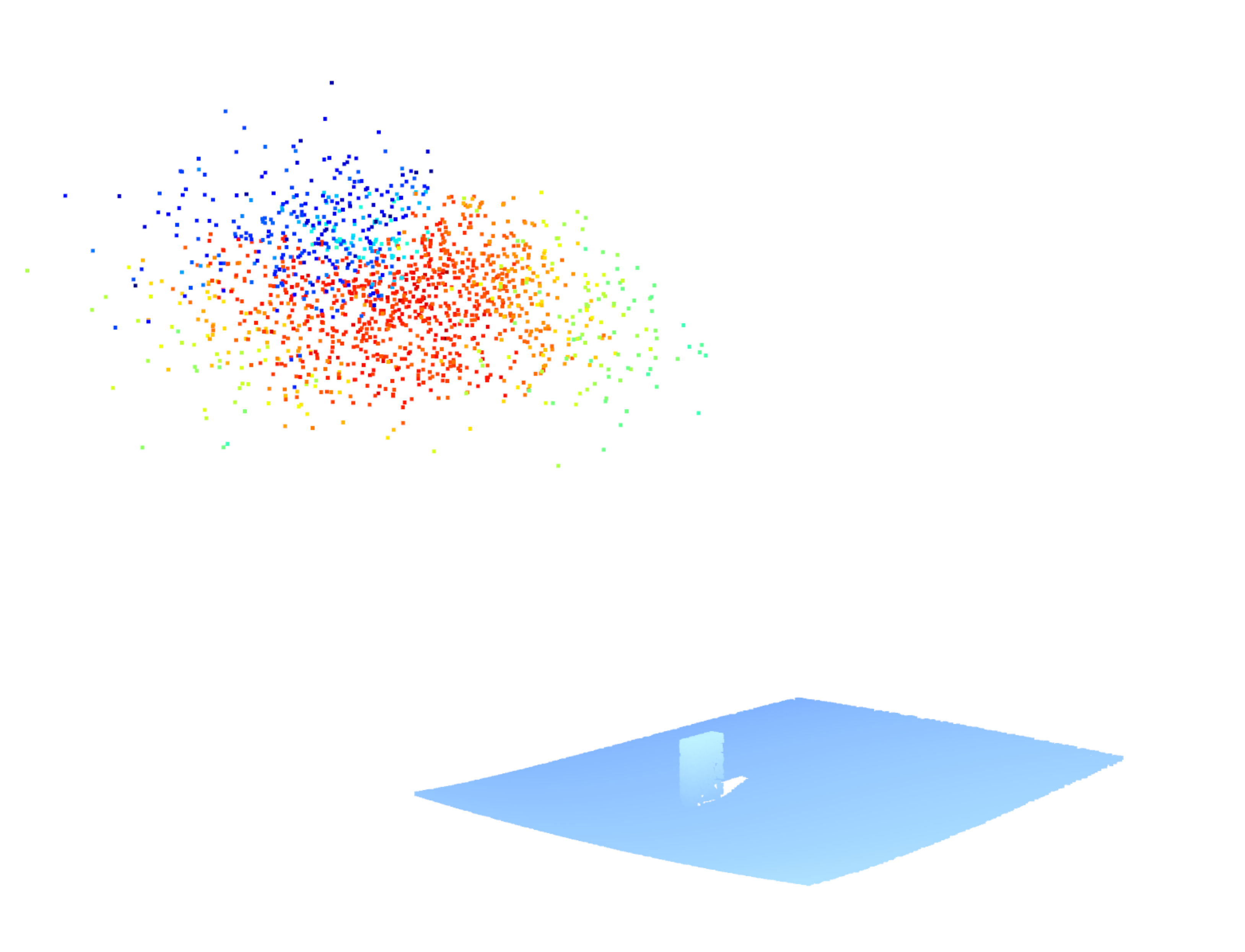}}
	\hfill
	\subfloat[]{\includegraphics[width=4.25cm]{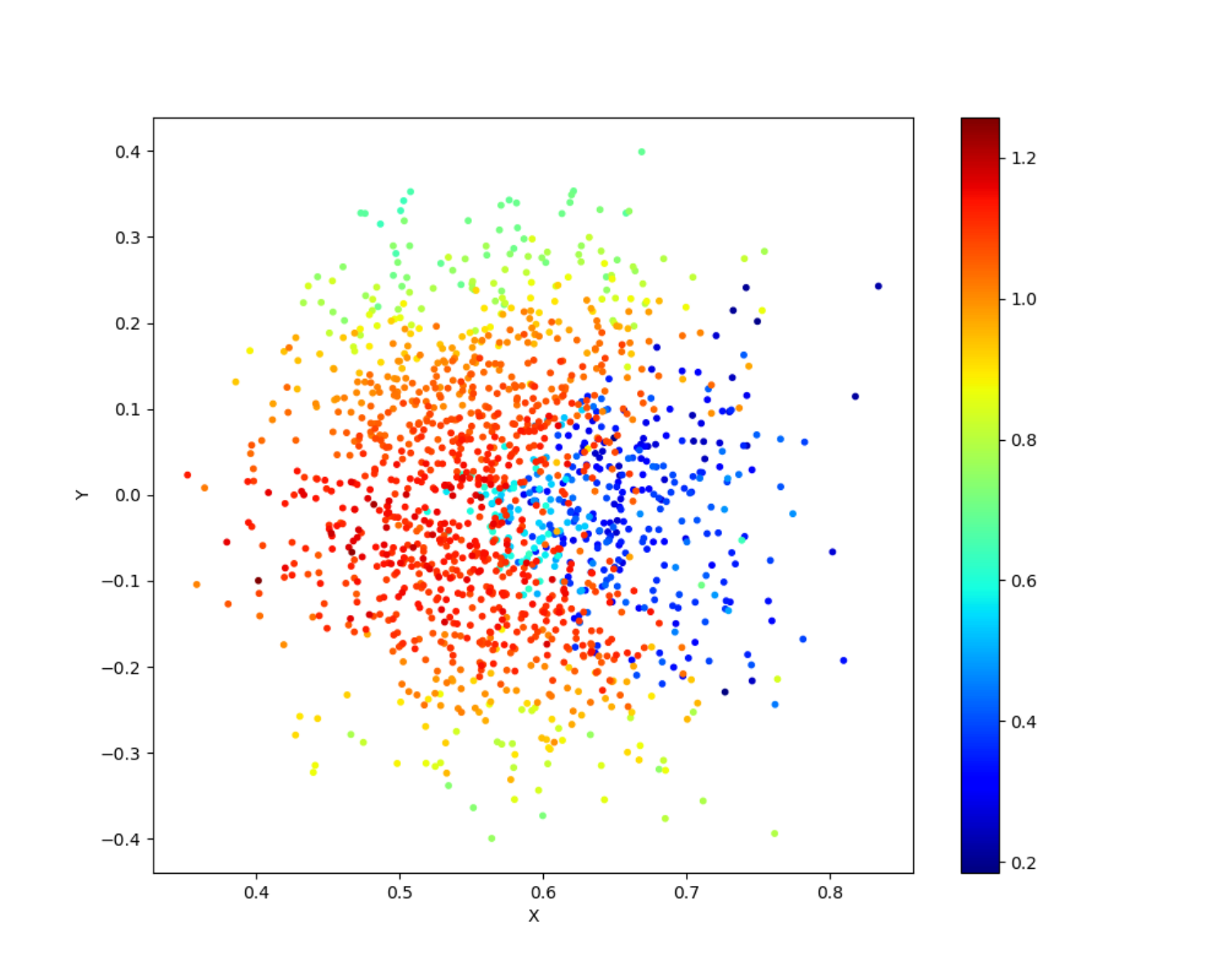}}
	\\
	\subfloat[]{\includegraphics[width=4.25cm]{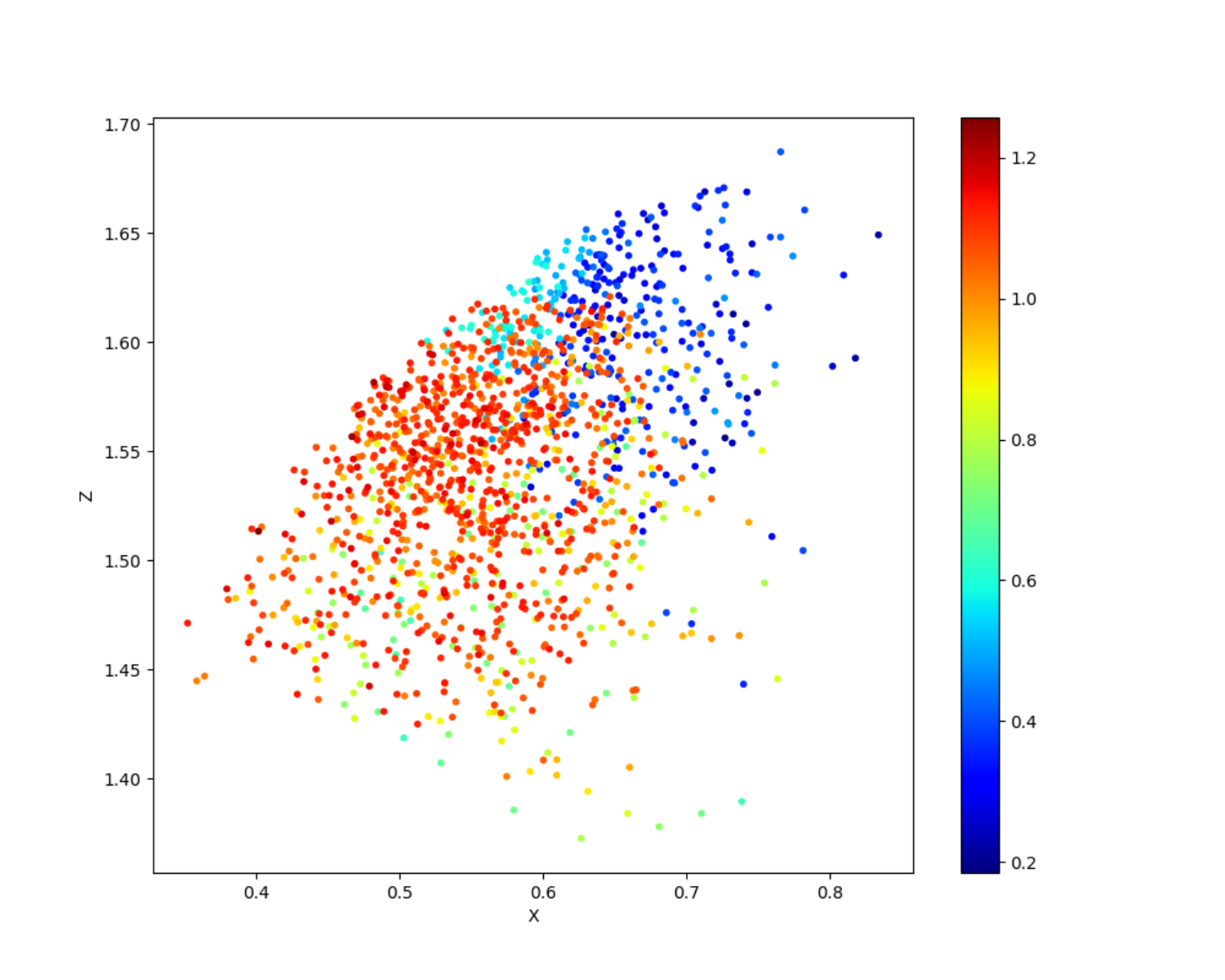}}
	\hfill
	\subfloat[]{\includegraphics[width=4.25cm]{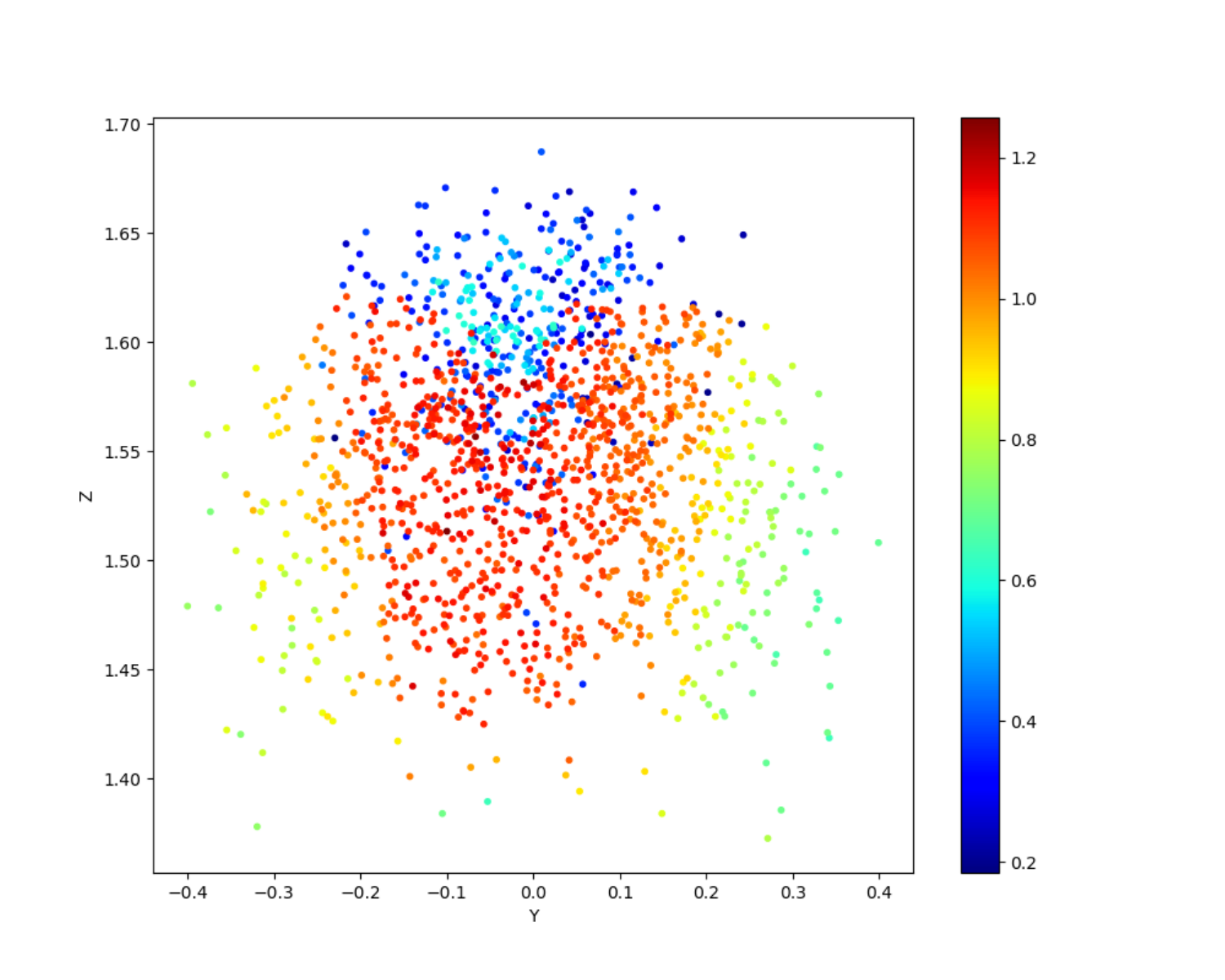}}
	\caption{An example of the viewpoint score distribution of the object: (a) 3D distribution; (b) X-Y plane projection; (c) X-Z plane projection; (d) Y-Z plane projection. Each point represents a observation viewpoint, with color indicating its score: \textcolor{red}{red} (highest), \textcolor{green}{green} (medium), and \textcolor{blue}{blue} (lowest).}
	\label{fig:synthetic_dataset_examples}
\end{figure}
\begin{figure*}[!t]\centering
	\includegraphics[width=17.1cm]{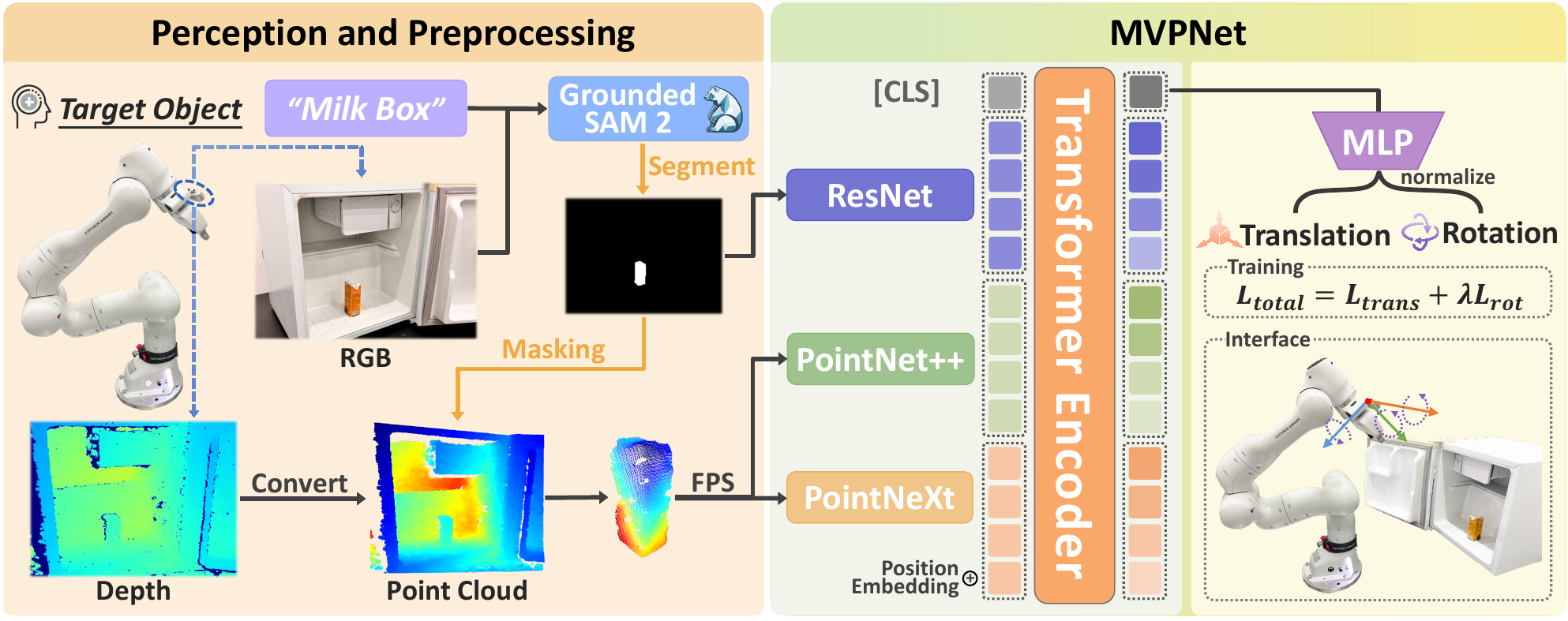}
	\caption{Overall framework of the perception and preprocessing modules together with MVPNet.}
	\label{fig:network}
\end{figure*}

The dataset consists of three components: mask images, point clouds, and camera pose adjustments.
The mask images and point clouds are obtained by preprocessing RGB-D observations in conjunction with the corresponding natural language instructions, and they serve as the inputs to the network, while the camera pose adjustments provide the learning targets.
Using the domain randomization capabilities in Isaac Sim, we collect 17k data samples. Subsequently, the dataset is split into training and testing sets following a 9:1 ratio.

\subsection{Perception and Preprocessing}
A single camera is utilized for image acquisition to facilitate policy learning, which enhances the applicability of our approach in real-world deployments \cite{Ze2024DP3}. For both simulation and real-world settings, color and depth images are acquired at a resolution of $1280 \times 720$ using a wrist-mounted camera, consistent with the dimensions used in the dataset. The depth image is then converted into a point cloud using the camera extrinsics and intrinsics.

Due to background interference, directly feeding raw camera-captured images into the network may be suboptimal. On the one hand, cluttered scenes and irrelevant objects make it difficult for the network to focus on the target object, thereby impeding network convergence. On the other hand, background geometry unrelated to the target object may introduce irrelevant features; for example, the model may inadvertently rely on the geometry of the tabletop when predicting camera pose adjustments.
Furthermore, in multi-object scenes, we aim to enable natural language-guided active perception for specific target objects. Therefore, the observations are preprocessed by incorporating target object language descriptions using Grounded SAM~2 \cite{ren2024grounded}, which combines Grounding DINO \cite{liu2023grounding} for object detection and SAM~2 \cite{ravi2024sam2segmentimages} for segmentation.
As shown in Fig. \mbox{\ref{fig:network}}, its workflow involves: 1) Grounding DINO detects the target object based on the semantic information of the object to be grasped and the camera-captured RGB image, and then passes its bounding box to SAM 2 for segmentation to generate a mask; 2) The mask serves two purposes: it is used directly as an image input to MVPNet and is also applied for point cloud masking; 3) The processed point cloud undergoes data augmentation (e.g., random point dropout) and farthest point sampling before being fed into MVPNet. Therefore, the inputs fed into the network are preprocessed and augmented, which not only facilitates the learning of salient features but also improves generalization to novel objects, unseen scenes, and even real-world environments.

\begin{remark}
Prior studies have shown that when images containing color, texture, and shape information are jointly provided as network input, convolutional neural networks tend to rely more on color and texture information than on geometric shape for prediction \cite{geirhos2018imagenet}. In this work, it is evident that the model should primarily focus on the boundary geometry of the target object to infer the required camera pose adjustment, instead of relying on color or texture information. Accordingly, the image input to the network is represented as a binary mask of the target object, rather than a color mask containing appearance information. This design not only helps the network learn key features but also further improves its generalization capability across variations in object color, texture, and environmental conditions (e.g., lighting).
\end{remark}

\subsection{Network Architecture Construction}
We propose a multimodal optimal observation viewpoint prediction network, termed MVPNet, which comprises three primary components: feature extraction, feature alignment and fusion, and pose output.

\subsubsection{Feature Extraction}\label{sec:feature_extraction}
We adopt ResNet \cite{He_2016_CVPR} to encode the mask image and use two point cloud encoders to extract point cloud features, as the point cloud contains richer three-dimensional spatial relationships compared to the mask image. This spatial data is more directly interpretable than image pixel values and resides in the same metric space as the output transformation $\Delta \mathbf{T}$.

Specifically, a combined point cloud encoder architecture integrating both PointNeXt \cite{qian2022pointnext} and PointNet++ \cite{qi2017pointnet++} is employed for collaborative feature extraction. Although PointNeXt is an improved version of PointNet++, each exhibits distinct characteristics in feature extraction: PointNet++ employs hierarchical downsampling with a focus on traditional geometric modeling, making it more sensitive to fine-grained structural details; PointNeXt incorporates enhanced residual connections optimized for modern deep feature extraction, providing superior awareness of global context and high-level semantics. By combining both architectures, our model effectively captures local geometric details while maintaining robust global representations, thereby enhancing adaptability across diverse input conditions.

\subsubsection{Feature Alignment and Fusion}
The feature extractors output modality-specific feature vectors, which are treated as tokens. These point cloud and mask image tokens are concatenated into a unified sequence and passed to a Transformer \cite{vaswani2017attention} encoder, whose cross-attention mechanism implicitly establishes 2D-3D correspondences and fuses their representations.
Moreover, a trainable classification ([CLS]) token is integrated within the Transformer encoder, similar to its use in BERT, where it serves as a summary representation for sequence-level prediction \cite{zhao2023learning}. Here, this token acts as a global feature vector that aggregates information from both point cloud and mask image tokens via the cross-attention mechanism, and its corresponding output is then employed to predict the required camera pose transformation. It ensures prediction robustness by integrating comprehensive geometric features from the point cloud with complementary visual silhouette information from the mask image, maintaining reliable performance even with incomplete observations or noisy inputs. The 2D sinusoidal position embedding is added to preserve spatial information \cite{carion2020end}.

\subsubsection{Pose Output}
We apply MLP to map the Transformer outputs to the task space, which consists of translation $\Delta \mathbf{t}$ and rotation $\Delta \mathbf{q}$. During training, these outputs are compared with ground-truth labels, and the network parameters are optimized through the loss function defined in \mbox{Section \ref{Loss Function}}. During inference, the predicted outputs are used to adjust the camera pose toward the optimal observation viewpoint.

\subsection{Loss Function}\label{Loss Function}
The loss function comprises two components: $\mathcal{L}_{trans}$ for translation and $\mathcal{L}_{rot}$ for rotation. Specifically, $\mathcal{L}_{trans}$ computes MSE between predicted translation $\hat{\mathbf{t}}$ and ground-truth translation $\mathbf{t}$, while $\mathcal{L}_{rot}$ computes the geodesic distance between predicted rotation $\hat{\mathbf{q}}$ and ground-truth rotation $\mathbf{q}$. The total loss is defined as
\begin{align}
\mathcal{L}_{total} = \mathcal{L}_{trans} + \lambda \mathcal{L}_{rot},
\end{align}
where
\begin{align}
\mathcal{L}_{trans} = \left\| \hat{\mathbf{t}} - \mathbf{t} \right\|_2^2, \;
\mathcal{L}_{rot} = 2 \cos^{-1} \left( \left| \left\langle \hat{\mathbf{q}}, \mathbf{q} \right\rangle \right| \right),
\label{eq:loss_function}
\end{align}
and $\lambda$ is a weighting parameter.
}

\section{Simulation and Evaluation}
Effectiveness and ablation experiments are first conducted in simulation on robotic grasping tasks under viewpoint-constrained environments to evaluate the proposed framework.

\subsection{Simulation Setup}
As shown in Fig.~\ref{fig:overall}\subref{fig:simutation_setup}, we construct a virtual environment based on the Isaac Sim 4.0.0\cite{website1} simulation platform, which includes a Franka Emika robotic arm equipped with its default parallel gripper. An Intel RealSense D435i camera mounted on the robotic arm wrist captures RGB-D images with a resolution of 1280 $\times$ 720. The camera's observation viewpoints are constrained such that top-down observations of the object are not available, and only side-view observations are allowed. The robotic arm serves dual purposes: executing grasping actions and acting as a carrier for the camera, enabling active perception through camera pose adjustment. The test involves 65 similar and 18 novel objects for grasping evaluation, as shown in Fig.~\ref{fig:object_similar} and Fig.~\ref{fig:object_novel}. Object categories, positions, scales, and orientations are randomly selected. The camera is placed at a random side position facing the object center, with random minor offsets in orientation to simulate realistic viewing conditions. It is worth noting that some objects within the similar objects category may be inherently more difficult to grasp than those in the novel objects category. As a result, the grasp success rate on similar objects can be lower than that on novel objects.

\begin{algorithm}
\caption{Simulation-Based Evaluation Pipeline}
\label{alg:simulation_evaluation}
\begin{algorithmic}[1]
\REQUIRE Total evaluation trials $n$
\ENSURE SR-1 and SR-5 for initial ($init$) and optimized ($opt$) viewpoints

\STATE \textbf{Initialize Accumulators:}
\STATE $Acc_{init}^{(1)} \leftarrow 0, \; Acc_{init}^{(5)} \leftarrow 0$; $Acc_{opt}^{(1)} \leftarrow 0, \; Acc_{opt}^{(5)} \leftarrow 0$

\FOR{$t = 1$ \TO $n$}
    \STATE \textbf{Setup:} Randomly initialize \textit{object}, \textit{camera}, \textit{env}; record state $S_0$

    \STATE \textit{// --- Phase 1: Evaluation at Initial Viewpoint ---}
    \STATE Acquire observation $\mathcal{O}_{init}$ and language $\mathcal{L}$
    \STATE Segment target object: $\mathcal{O}'_{init} = M_s(\mathcal{O}_{init}, \mathcal{L})$
    
    \STATE $(c_1, c_5) \leftarrow \textsc{GraspExecution}(\mathcal{O}'_{init}, S_0)$
    
    \STATE $Acc_{init}^{(1)} \leftarrow Acc_{init}^{(1)} + c_1$, $Acc_{init}^{(5)} \leftarrow Acc_{init}^{(5)} + c_5$

    \STATE \textit{// --- Phase 2: Evaluation at Optimized Viewpoint ---}
    \STATE Predict camera pose adjustment: $\Delta \mathbf{T} = M_v(\mathcal{O}'_{init})$
    \STATE Move camera according to $\Delta \mathbf{T}$
    \STATE Acquire optimized observation $\mathcal{O}_{opt}$
    \STATE Segment target object: $\mathcal{O}'_{opt} = M_s(\mathcal{O}_{opt}, \mathcal{L})$
    
    \STATE $(c_1, c_5) \leftarrow \textsc{GraspExecution}(\mathcal{O}'_{opt}, S_0)$
    
    \STATE $Acc_{opt}^{(1)} \leftarrow Acc_{opt}^{(1)} + c_1$, $Acc_{opt}^{(5)} \leftarrow Acc_{opt}^{(5)} + c_5$
\ENDFOR

\STATE \textbf{Compute Final Metrics:}
\STATE $\text{SR-1}_{init} = Acc_{init}^{(1)} / n, \; \text{SR-5}_{init} = Acc_{init}^{(5)} / (5 \times n)$

\STATE $\text{SR-1}_{opt} = Acc_{opt}^{(1)} / n, \; \text{SR-5}_{opt} = Acc_{opt}^{(5)} / (5 \times n)$

\vspace{0.5em} 
\STATE \hrulefill 
\vspace{0.5em}

\STATE \textbf{Function} \textsc{GraspExecution}($\mathcal{O}', S_{\text{reset}}$)
    \STATE Estimate grasp candidates: $\mathcal{P} = M_g(\mathcal{O}')$
    \STATE Select top-5 candidates: $\mathcal{C} = \text{Top-5}(\mathcal{P})$
    
    \STATE $count_1 \leftarrow 0, \quad count_5 \leftarrow 0$
    
    \FOR{$k = 1$ \TO $5$}
        \STATE Execute candidate $c_k \in \mathcal{C}$ in simulation
        \IF{lift is successful}
            \STATE $count_5 \leftarrow count_5 + 1$
            \IF{$k=1$} 
                \STATE $count_1 \leftarrow 1$ 
            \ENDIF
        \ENDIF
        \STATE Reset environment to state $S_{\text{reset}}$
    \ENDFOR
    \RETURN $(count_1, count_5)$
\end{algorithmic}
\end{algorithm}

\subsection{Implementation Details}
Each experimental group undergoes 250 trials with the following evaluation protocol: First, we randomly initialize the camera and object poses, and then collect RGB-D observations from the current viewpoint. The RGB image, together with the semantic information of the target object, is provided to Grounded-SAM2\cite{ren2024grounded} to obtain an instance mask, which is then applied to the point cloud to isolate the object of interest. The masked observations are subsequently fed into the grasping model to estimate grasp poses. We then select the top-5 highest-scoring grasp candidates and execute them sequentially in Isaac Sim to evaluate the grasping performance at the initial viewpoint. Subsequently, MVPNet predicts the optimal observation viewpoint, after which the camera is moved to that viewpoint to acquire new observations. The segmentation model and the grasping model are applied again for grasp pose estimation, followed by the same candidate selection and execution process. The overall procedure is summarized in Procedure~\ref{alg:simulation_evaluation}. We record both the first-attempt success rate (SR-1) and the average success rate over five attempts (SR-5) for comprehensive analysis. All success rates (SR-1 and SR-5) are reported in percentage (\%). Grasping success is measured by lifting the object 10 cm without dropping it.

We train the model using the AdamW \cite{loshchilov2017decoupled} optimizer with a weight decay of $1 \times 10^{-4}$. The initial learning rate for the main network is set to $5 \times 10^{-5}$, while the backbone, a pre-trained ResNet-18, uses a lower rate of $5 \times 10^{-6}$. A StepLR scheduler is adopted, in which the learning rate decays by a factor of 0.7 every 20 epochs. The model is trained for 100 epochs with a batch size of 16. The weighting parameter $\lambda$ in Eq.~\mbox{(\ref{eq:loss_function})} is set to 1. All training and inference are conducted on a workstation equipped with an Intel Xeon Silver 4416+ CPU and a single NVIDIA RTX 4090 GPU.
\subsection{Effectiveness}

Four representative grasp pose estimation models are used in the experiments: Economic Grasp \cite{wu2024economic}, GraspNet \cite{fang2020graspnet}, TRG \cite{yu2025trustworthy}, and GIGA \cite{jiang2021synergies}. Economic Grasp and GraspNet take point clouds as input, whereas TRG and GIGA use TSDF as input.
This setup enables us to directly evaluate the performance improvement our active perception brings to various types of grasp estimation pipelines.

The results in Tables~\ref{table_simulation_experiments_all} show that, after optimizing the initial observation viewpoints using MVPNet, the grasp success rates of all baselines are significantly improved for both similar objects and novel objects. This improvement is observed even though the observation viewpoint quality evaluation function is based on the Economic Grasp metric, indicating that the proposed framework is practical across different grasp pose estimation models. Specifically, for point cloud-based grasp models such as Economic Grasp and GraspNet, our approach results in clear gains in both SR-5 and SR-1, indicating that a more informative view helps these models generate more reliable grasp proposals. Compared with Economic Grasp, the improvement on GraspNet is more pronounced, as GraspNet is more sensitive to missing or incomplete geometric observations, and thus benefits more from improved viewpoints. TSDF-based models such as TRG and GIGA also benefit from active perception.
It should be noted that TRG and GIGA generate relatively few grasp poses; therefore, only SR-1 is reported. Furthermore, since they are primarily designed for grasp pose estimation from a fixed view, their grasp success rates under dynamic views are comparatively low.

Overall, these results demonstrate that selecting more informative viewpoints guided by MVPNet provides grasp models with richer and more discriminative observations, enabling more accurate grasp pose estimations, and further validate the effectiveness of the proposed framework.

\begin{table}[!t]
\caption{Simulation Results of Different Grasp Pose Estimation Models Using Initial and Optimized Views.}
\centering
\label{table_simulation_experiments_all}
\setlength{\tabcolsep}{3pt}
\begin{tabularx}{8.8cm}{l c|>{\centering\arraybackslash}X
                         >{\centering\arraybackslash}X|
                         >{\centering\arraybackslash}X
                         >{\centering\arraybackslash}X}
\toprule[1pt]
\multirow{2}{*}{Model} & \multirow{2}{*}{Viewpoint}
& \multicolumn{2}{c|}{Similar Objects}
& \multicolumn{2}{c}{Novel Objects} \\
\cmidrule{3-6}
& & SR-5 & SR-1 & SR-5 & SR-1 \\
\midrule
\multirow{2}{*}{\makecell[l]{Economic\\Grasp}}
& \textbf{Optimized (Ours)}
& \cellcolor[HTML]{CCF2F5}\textbf{64.8}
& \cellcolor[HTML]{CCF2F5}\textbf{64.0}
& \cellcolor[HTML]{CCF2F5}\textbf{63.2}
& \cellcolor[HTML]{CCF2F5}\textbf{64.7} \\
& Initial
& 54.4 & 53.4 & 49.2 & 48.0 \\
\midrule
\multirow{2}{*}{GraspNet}
& \textbf{Optimized (Ours)}
& \cellcolor[HTML]{CCF2F5}\textbf{49.6}
& \cellcolor[HTML]{CCF2F5}\textbf{54.0}
& \cellcolor[HTML]{CCF2F5}\textbf{51.3}
& \cellcolor[HTML]{CCF2F5}\textbf{55.3} \\
& Initial
& 32.0 & 32.7 & 32.3 & 50.7 \\
\midrule
\multirow{2}{*}{TRG}
& \textbf{Optimized (Ours)}
& -- & \cellcolor[HTML]{CCF2F5}\textbf{24.0}
& -- & \cellcolor[HTML]{CCF2F5}\textbf{35.6} \\
& Initial
& -- & 22.0 & -- & 32.0 \\
\midrule
\multirow{2}{*}{GIGA}
& \textbf{Optimized (Ours)}
& -- & \cellcolor[HTML]{CCF2F5}\textbf{26.0}
& -- & \cellcolor[HTML]{CCF2F5}\textbf{28.0} \\
& Initial
& -- & 19.3 & -- & 23.3 \\
\midrule
\multicolumn{2}{l|}{\textbf{Average Improvement}}
& +14.0 & +10.2 & +16.5 & +7.4 \\
\bottomrule[1pt]
\end{tabularx}
\end{table}

\subsection{Ablations}
In this subsection, detailed ablation studies are conducted to validate the effectiveness of the MVPNet design, covering input modalities, point cloud encoders, and network architecture. The ablation studies employ the same experimental setup and evaluation metrics as above, using Economic Grasp and GraspNet as the grasping models.
\subsubsection{Choice of Input Modalities}
Point cloud and mask image provide complementary 3D and 2D representations of the object, respectively. Leveraging both modalities enables the network to capture object geometry and silhouettes more comprehensively, leading to more reliable prediction. As shown in \mbox{Table \ref{table_ablation_input_modalities_economic}} and \mbox{\ref{table_ablation_input_modalities_graspnet}}, the multimodal configuration that fuses point cloud and mask image achieves the best performance across most metrics, with especially strong gains on the novel objects SR-5 metric, which reflects comprehensive generalization capability. Consequently, removing either modality consistently degrades performance, highlighting the importance of multimodal input for robust optimal viewpoint prediction.
\begin{table}[!h]
	\caption{Ablation on Input Modalities (Economic Grasp).}
	\centering
	\label{table_ablation_input_modalities_economic}
	\begin{tabularx}{0.5\textwidth}{l|>{\centering\arraybackslash}X|
		>{\centering\arraybackslash}X|
		>{\centering\arraybackslash}X|
		>{\centering\arraybackslash}X}
	\toprule[1pt]
	\midrule
	 & \multicolumn{2}{c|}{Similar Objects} & \multicolumn{2}{c}{Novel Objects} \\ \cmidrule{2-5} 
	\multirow{-2}{*}{Input Modalities} & \multicolumn{1}{c|}{SR-5} & SR-1 & \multicolumn{1}{c|}{SR-5} & SR-1 \\ \midrule
	\textbf{Point Cloud + Mask (Ours)} & \cellcolor[HTML]{CCF2F5}\textbf{66.8} & \cellcolor[HTML]{CCF2F5}\textbf{64.0} & \cellcolor[HTML]{CCF2F5}\textbf{62.8} & \cellcolor[HTML]{CCF2F5}\textbf{60.0} \\
	w/o Point Cloud & 59.2 & 62.0 & 58.4 & 58.0 \\ 
	w/o Mask & 64.8 & \cellcolor[HTML]{CCF2F5}\textbf{64.0} & 49.6 & 48.0 \\ 
	\bottomrule[1pt]
	\end{tabularx}
	\end{table}
\begin{table}[!h]
	\caption{Ablation on Input Modalities (GraspNet).}
	\centering
	\label{table_ablation_input_modalities_graspnet}
	\begin{tabularx}{0.5\textwidth}{l|>{\centering\arraybackslash}X|
		>{\centering\arraybackslash}X|
		>{\centering\arraybackslash}X|
		>{\centering\arraybackslash}X}
	\toprule[1pt]
	\midrule
	 & \multicolumn{2}{c|}{Similar Objects} & \multicolumn{2}{c}{Novel Objects} \\ \cmidrule{2-5} 
	\multirow{-2}{*}{Input Modalities} & \multicolumn{1}{c|}{SR-5} & SR-1 & \multicolumn{1}{c|}{SR-5} & SR-1 \\ \midrule
	\textbf{Point Cloud + Mask (Ours)} & 53.2 & 54.0 & \cellcolor[HTML]{CCF2F5}\textbf{42.8} & 46.0 \\
	w/o Point Cloud & 49.6 & 54.0 & 41.2 & 48.0 \\ 
	w/o Mask & \cellcolor[HTML]{CCF2F5}\textbf{58.0} & \cellcolor[HTML]{CCF2F5}\textbf{60.0} & 40.4 & \cellcolor[HTML]{CCF2F5}\textbf{52.0} \\ 
	\bottomrule[1pt]
	\end{tabularx}
	\end{table}

\subsubsection{Choice of Point Cloud Encoders}\label{sec:choice_pointcloud_encoders}
We compare the point cloud encoders in MVPNet, which integrates PointNeXt \cite{qian2022pointnext} and PointNet++ \cite{qi2017pointnet++}, with several widely adopted alternatives, including Point Transformer \cite{wu2024ptv3} and LayerNorm \cite{ba2016layer} (demonstrated to be effective in DP3 \cite{Ze2024DP3}).
The results in \mbox{Table \ref{table_ablation_pc_encoder_economic}} and \mbox{Table \ref{table_ablation_pc_encoder_graspnet}} demonstrate that the combination of PointNeXt and PointNet++ consistently achieves the best performance across all metrics, confirming the effectiveness of jointly leveraging both encoders.
As mentioned in Section \ref{sec:feature_extraction}, PointNeXt captures global geometric structures, while PointNet++ focuses on local feature extraction, making their fusion highly complementary.
The combination improves feature extraction robustness across different object categories, initial viewpoints, and sensing conditions, leading to more stable viewpoint prediction performance.

\begin{table}[!h]
\caption{Ablation on Point Cloud Encoders (Economic Grasp).}
\centering
\label{table_ablation_pc_encoder_economic}
\begin{tabularx}{0.5\textwidth}{l|>{\centering\arraybackslash}X|
	>{\centering\arraybackslash}X|
	>{\centering\arraybackslash}X|
	>{\centering\arraybackslash}X}
\toprule[1pt]
\midrule
                                    & \multicolumn{2}{c|}{Similar Objects}                                      & \multicolumn{2}{c}{Novel Objects}                                         \\ \cmidrule{2-5} 
\multirow{-2}{*}{Point Cloud Encoders}          & \multicolumn{1}{c|}{SR-5}           & SR-1                                & \multicolumn{1}{c|}{SR-5}           & SR-1                                \\ \midrule
\textbf{PointNeXt + PointNet++ (Ours)} & \cellcolor[HTML]{CCF2F5}\textbf{66.8} & \cellcolor[HTML]{CCF2F5}\textbf{64.0} & \cellcolor[HTML]{CCF2F5}\textbf{62.4} & \cellcolor[HTML]{CCF2F5}\textbf{68.0} \\
PointNet++ + LayerNorm                & 62.4                                & 62.0                                  & 51.6                                & 52.0                                  \\
PointNeXt + LayerNorm                 & 60.4                                & \cellcolor[HTML]{CCF2F5}\textbf{64.0} & 51.6                                & 46.0                                  \\
PointNeXt + Point Transformer         & 65.2                                & \cellcolor[HTML]{CCF2F5}\textbf{64.0} & 59.6                                & 54.0                                  \\ 
Point Transformer + LayerNorm         & 54.0                                & 52.0                                  & 58.4                                & 60.0                                  \\
Point Transformer + PointNet++        & 59.6                                & \cellcolor[HTML]{CCF2F5}\textbf{64.0} & 43.2                                & 46.0                                  \\ \midrule
\bottomrule[1pt]
\end{tabularx}
\end{table}

\begin{table}[!h]
\caption{Ablation on Point Cloud Encoders (GraspNet).}
\centering
\label{table_ablation_pc_encoder_graspnet}
\begin{tabularx}{0.5\textwidth}{l|>{\centering\arraybackslash}X|
	>{\centering\arraybackslash}X|
	>{\centering\arraybackslash}X|
	>{\centering\arraybackslash}X}
\toprule[1pt]
\midrule
                                    & \multicolumn{2}{c|}{Similar Objects}                                      & \multicolumn{2}{c}{Novel Objects}                                         \\ \cmidrule{2-5} 
\multirow{-2}{*}{Point Cloud Encoders}          & \multicolumn{1}{c|}{SR-5}           & SR-1                                & \multicolumn{1}{c|}{SR-5}           & SR-1                                \\ \midrule
\textbf{PointNeXt + PointNet++ (Ours)} 
    & \cellcolor[HTML]{CCF2F5}\textbf{46.8} 
    & \cellcolor[HTML]{CCF2F5}\textbf{50.0}
    & \cellcolor[HTML]{CCF2F5}\textbf{60.0} 
    & \cellcolor[HTML]{CCF2F5}\textbf{68.0} \\
PointNet++ + LayerNorm          & 40.0 & 40.0 & 55.6 & 64.0 \\
PointNeXt + LayerNorm           & 38.4 & 44.0 & 54.8 & 62.0 \\
PointNeXt + Point Transformer   & 42.0 & 46.0 & 52.0 & 60.0 \\ 
Point Transformer + LayerNorm   & 41.2 & 48.0 & 50.4 & 62.0 \\
Point Transformer + PointNet++  & 34.8 & 42.0 & 50.4 & 58.0 \\ 
\bottomrule[1pt]
\end{tabularx}
\end{table}

\subsubsection{Choice of Network Architecture}
We conduct network architectural ablation studies by individually removing PointNet++ and PointNeXt and replacing the Transformer with an MLP. As shown in \mbox{Table \ref{table_ablation_network_architecture_economic}} and \mbox{Table \ref{table_ablation_network_architecture_graspnet}}, and as mentioned in Section \ref{sec:feature_extraction} and \ref{sec:choice_pointcloud_encoders}, PointNet++ and PointNeXt extract point cloud features at different scales, and this multi-scale representation enhances overall network performance. Simultaneously, the Transformer offers superior capability for multimodal alignment and information fusion between point cloud and mask image features.
\begin{table}[!h]
	\caption{Ablation on Network Architecture (Economic Grasp).}
	\centering
	\label{table_ablation_network_architecture_economic}
	\begin{tabularx}{0.5\textwidth}{l|>{\centering\arraybackslash}X|
		>{\centering\arraybackslash}X|
		>{\centering\arraybackslash}X|
		>{\centering\arraybackslash}X}
	\toprule[1pt]
	\midrule
	 & \multicolumn{2}{c|}{Similar Objects} & \multicolumn{2}{c}{Novel Objects} \\ \cmidrule{2-5} 
	\multirow{-2}{*}{Design Choices} & \multicolumn{1}{c|}{SR-5} & SR-1 & \multicolumn{1}{c|}{SR-5} & SR-1 \\ \midrule
	\textbf{MVPNet (Ours)} & 60.8 & \cellcolor[HTML]{CCF2F5}\textbf{64.0} & \cellcolor[HTML]{CCF2F5}\textbf{64.4} & \cellcolor[HTML]{CCF2F5}\textbf{66.0} \\
	w/o PointNet++ & \cellcolor[HTML]{CCF2F5}\textbf{62.4} & 62.0 & 60.0 & 60.0 \\ 
	w/o PointNeXt & 51.6 & 52.0 & 54.4 & 60.0 \\ 
	w/o Transformer & 50.4 & 52.0 & 57.6 & 62.0 \\ 
	\bottomrule[1pt]
	\end{tabularx}
	\end{table}

\begin{table}[!h]
	\caption{Ablation on Network Architecture (GraspNet).}
	\centering
	\label{table_ablation_network_architecture_graspnet}
	\begin{tabularx}{0.5\textwidth}{l|>{\centering\arraybackslash}X|
		>{\centering\arraybackslash}X|
		>{\centering\arraybackslash}X|
		>{\centering\arraybackslash}X}
	\toprule[1pt]
	\midrule
	 & \multicolumn{2}{c|}{Similar Objects} & \multicolumn{2}{c}{Novel Objects} \\ \cmidrule{2-5} 
	\multirow{-2}{*}{Design Choices} & \multicolumn{1}{c|}{SR-5} & SR-1 & \multicolumn{1}{c|}{SR-5} & SR-1 \\ \midrule
	\textbf{MVPNet (Ours)} & \cellcolor[HTML]{CCF2F5}\textbf{48.8} & \cellcolor[HTML]{CCF2F5}\textbf{58.0} & \cellcolor[HTML]{CCF2F5}\textbf{51.2} & 52.0 \\
	w/o PointNet++ & 46.0 & 52.0 & 48.8 & 52.0 \\ 
	w/o PointNeXt & 46.4 & 50.0 & 48.4 & \cellcolor[HTML]{CCF2F5}\textbf{56.0} \\ 
	w/o Transformer & 48.4 & 56.0 & 46.0 & \cellcolor[HTML]{CCF2F5}\textbf{56.0} \\ 
	\bottomrule[1pt]
	\end{tabularx}
	\end{table}

Overall, although certain variants achieve slightly higher scores on individual metrics, MVPNet delivers the most consistent and robust performance across the majority of settings, particularly in SR-5 on Novel Objects. This demonstrates the necessity and complementarity of all architectural components and design choices.

\section{Real World Experiments}
\begin{figure}[!h]\centering
    \begin{subfigure}[b]{0.48\linewidth}
        \centering
        \includegraphics[width=\linewidth]{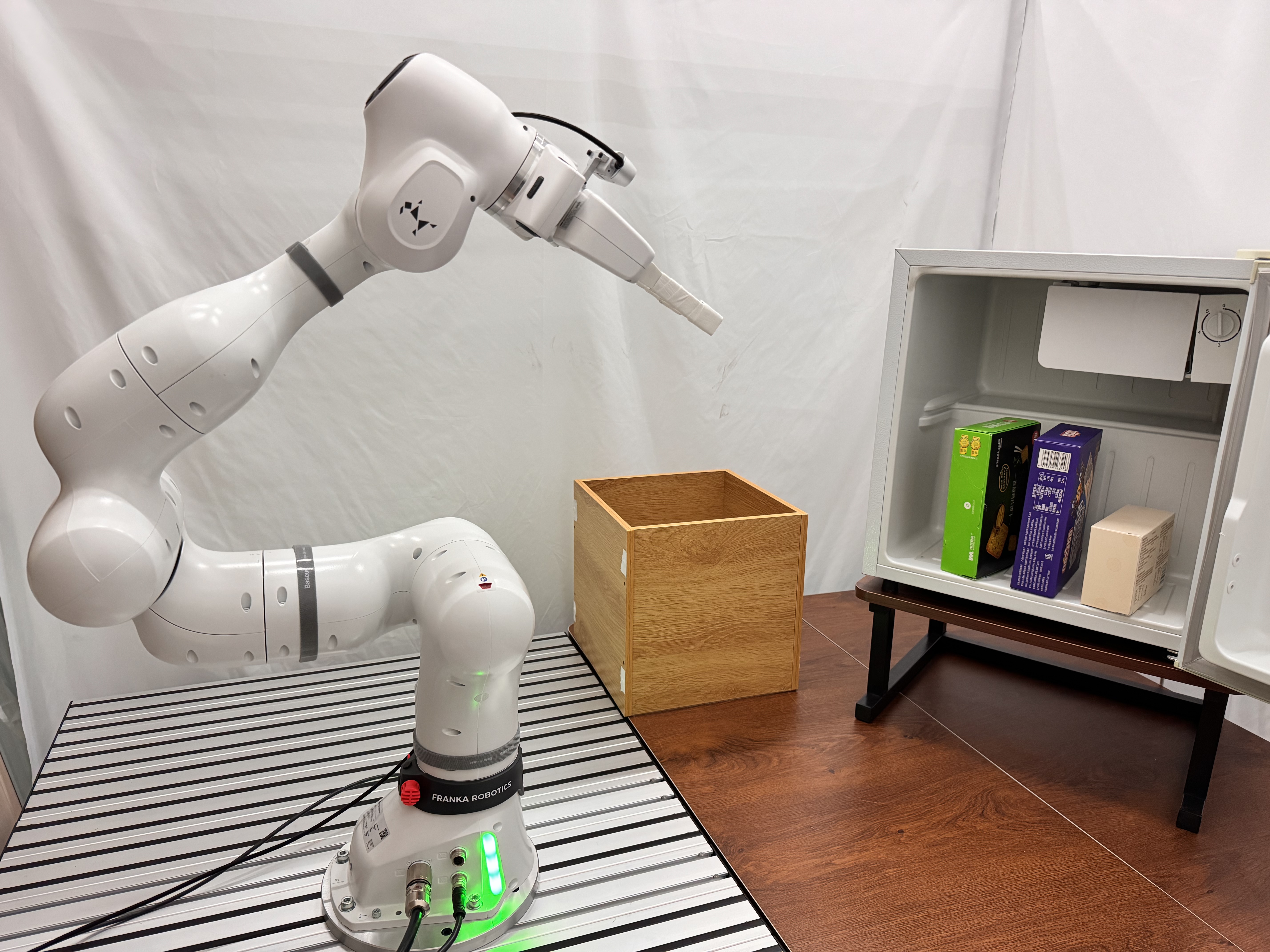}
        \caption{}
        \label{fig:real_world_setup_scene}
    \end{subfigure}
    \hfill
    \begin{subfigure}[b]{0.48\linewidth}
        \centering
        \includegraphics[width=\linewidth]{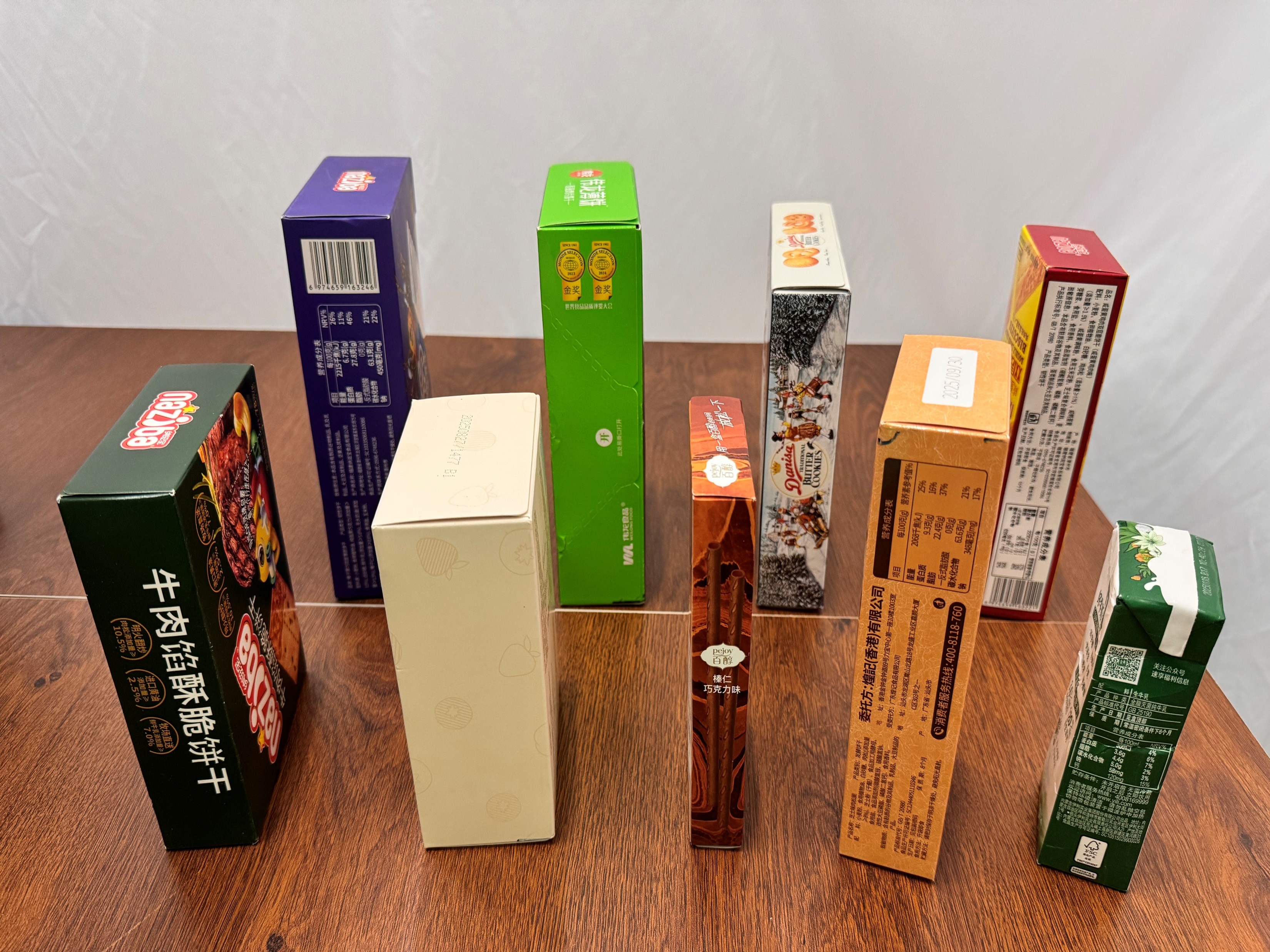}
        \caption{}
        \label{fig:real_world_setup_objects}
    \end{subfigure}
    \caption{(a) Real-world experimental scene setup; (b) Experimental objects.}
    \label{fig:real_world_setup}
\end{figure}

The effectiveness of the proposed framework is further validated through real-world experiments, with the experimental setup and tested objects shown in Fig.~\ref{fig:real_world_setup_scene} and Fig.~\ref{fig:real_world_setup_objects}, respectively. An Intel RealSense D435 camera is used to capture visual data, and a Franka Research 3 robotic arm is employed both to execute object grasping and to reposition the camera for active perception. We conduct 20 evaluation rounds, in which three objects are randomly selected and placed for each round. In every round, we first execute grasping without optimal viewpoint prediction, where grasp poses are predicted directly from the initial camera viewpoint. We then apply active perception independently to each of the three objects, move the camera to the corresponding optimized viewpoints, and repeat grasp pose prediction and execution. An object is considered successfully grasped if it is picked up and placed into the bin, whereas two consecutive failed grasp attempts are counted as an object grasp failure. We report two evaluation metrics: (1) Grasp Success Rate (GSR), defined as the ratio of successful grasp executions; and (2) Declutter Rate (DR), defined as the average fraction of objects removed. Their formal definitions are given as
\begin{align}
\mathrm{GSR} = \frac{n}{N}, \;
\mathrm{DR} = \frac{n_c}{N_{\mathrm{obj}}},
\end{align}
where \(n\) denotes the number of successful grasp attempts out of \(N\) total attempts, and \(n_c\) denotes the number of removed objects out of \(N_{\mathrm{obj}}\) total objects.

\begin{figure}[h]\centering
	\includegraphics[width=8.8cm]{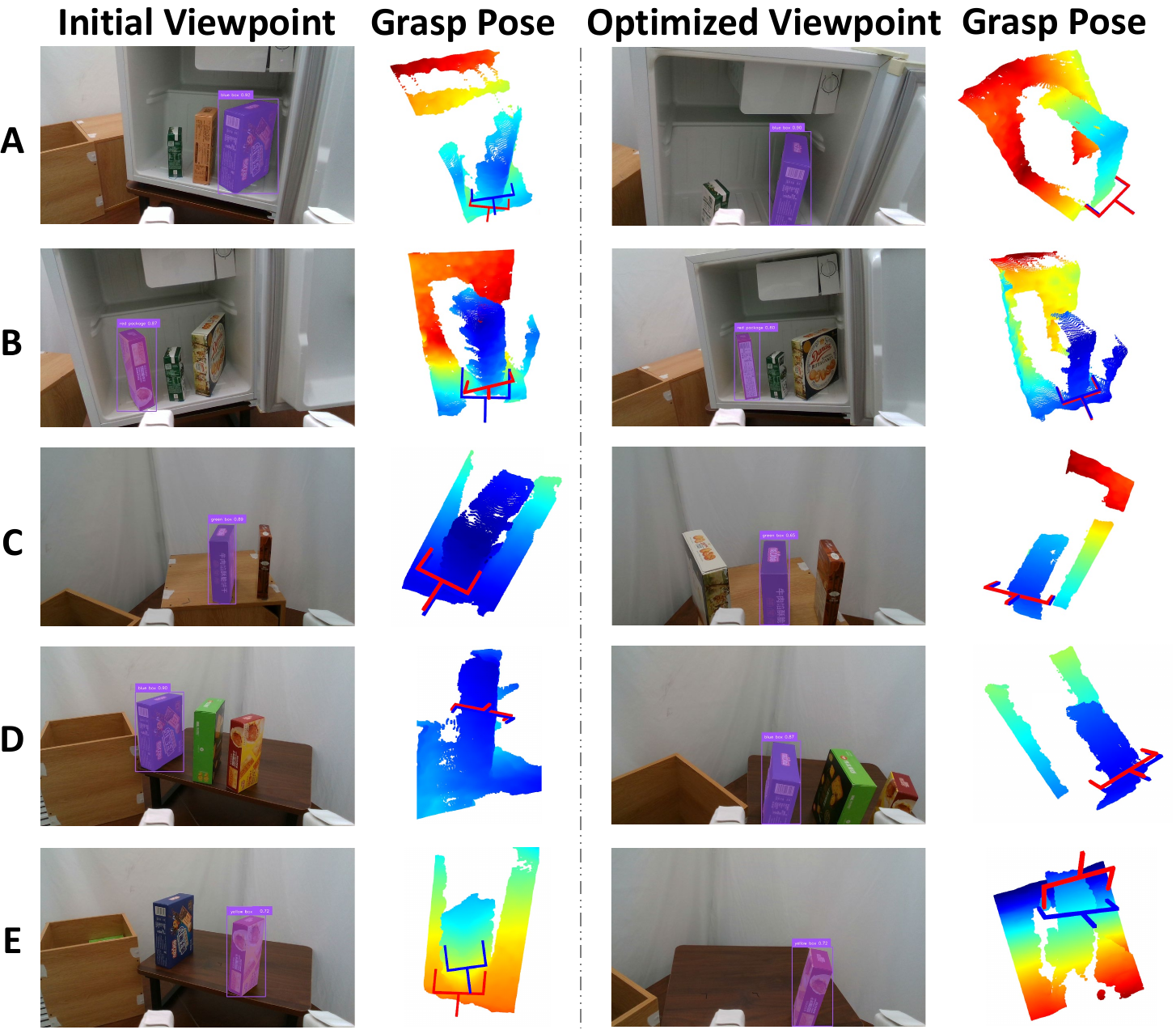}
	\caption{Active perception in real-world scenarios. The second and fourth columns present the grasp poses generated by Economic Grasp under the initial and optimized viewpoints, respectively. \textcolor{red}{The red gripper} indicates the highest-scoring grasp pose, while \textcolor{blue}{the blue gripper} denotes the second-best candidate. The target object to be grasped is indicated by the bounding box and the masked region.}
	\label{fig:real_world_demo}
\end{figure}

Table \ref{table:real_world_results} summarizes the real-world performance. After applying active perception using MVPNet, both the grasp success rate and declutter rate exhibit substantial improvements, with the grasp success rate nearly doubling. Fig. \ref{fig:real_world_demo} presents several representative examples, showing that the generated grasp poses become more accurate and reasonable after active perception. It is noteworthy that our model is trained entirely on synthetic data, without any real-world fine-tuning, demonstrating seamless sim-to-real transfer.  The real-world experimental results further demonstrate the effectiveness of the proposed framework.

\begin{table}[h]
    \caption{Real-World Results of Economic Grasp Using Initial and Optimized View.}
    \centering
    \label{table:real_world_results}
    \begin{tabularx}{0.48\textwidth}{l|>{\centering\arraybackslash}X|>{\centering\arraybackslash}X}
    \toprule[1pt]
    \textbf{Grasp Viewpoint} & \textbf{GSR (\%)} & \textbf{DR (\%)} \\ 
    \midrule
    \textbf{Optimized (Ours)} & \cellcolor[HTML]{CCF2F5}\textbf{47.6 (40/84)} & \cellcolor[HTML]{CCF2F5}\textbf{66.7} \\
    Initial & 25.5 (26/102) & 43.3 \\
    \bottomrule[1pt]
    \end{tabularx}
\end{table}

\section{Conclusion}

This paper aims to present a general active perception framework capable of optimal viewpoint prediction in a one-shot manner, comprising a data collection pipeline and an optimal viewpoint prediction network. The effectiveness of the proposed framework is validated through robotic grasping tasks in viewpoint-constrained environments, evaluated in both simulation and real-world settings. Moreover, the proposed framework enables seamless sim-to-real transfer, facilitating large-scale data collection in simulation and direct deployment in real-world systems without additional fine-tuning.

Future work will focus on extending the proposed framework to a broader range of robotic tasks, as well as exploring richer viewpoint representations to support more complex and dynamic manipulation scenarios.

\bibliographystyle{Bibliography/IEEEtranTIE}
\bibliography{Bibliography/IEEEabrv,Bibliography/BIB_xx-TIE-xxxx}\ 

\end{document}